\def\year{2022}\relax
\documentclass[letterpaper]{article} % DO NOT CHANGE THIS
\usepackage{aaai22}  % DO NOT CHANGE THIS
\usepackage{times}  % DO NOT CHANGE THIS
\usepackage{helvet}  % DO NOT CHANGE THIS
\usepackage{courier}  % DO NOT CHANGE THIS
\usepackage[hyphens]{url}  % DO NOT CHANGE THIS
\usepackage{graphicx} % DO NOT CHANGE THIS
\usepackage{natbib}  % DO NOT CHANGE THIS AND DO NOT ADD ANY OPTIONS TO IT
\usepackage{caption} % DO NOT CHANGE THIS AND DO NOT ADD ANY OPTIONS TO IT
\DeclareCaptionStyle{ruled}{labelfont=normalfont,labelsep=colon,strut=off} % DO NOT CHANGE THIS
\usepackage{algorithm}
\usepackage{algorithmic}
\usepackage{amsmath}
\usepackage{subfigure}
\usepackage{multirow}
\usepackage{bbding}
\usepackage{amssymb}% http://ctan.org/pkg/amssymb
\usepackage{pifont}% http://ctan.org/pkg/pifont
\usepackage{arydshln}
\usepackage{color}

\usepackage{newfloat}
\usepackage{listings}
\floatstyle{ruled}
\newfloat{listing}{tb}{lst}{}
\floatname{listing}{Listing}
\title{Frequency-Aware Contrastive Learning for Neural Machine Translation}

\author{Tong Zhang$^{1}$
Wei Ye$^{1,}$\footnote{Corresponding authors.}, Baosong Yang$^{2}$, Long Zhang$^{1}$, Xingzhang Ren$^{2}$,\\
\bf{Dayiheng Liu$^{2}$, Jinan Sun$^{1,}$\footnotemark[1], Shikun Zhang$^{1}$, Haibo Zhang$^{2}$, Wen Zhao$^{1}$}\\
\textmd{$^1$ National Engineering Research Center for Software Engineering, Peking University} \\ 
\textmd{$^2$ Alibaba Group}\\
\textmd{\normalsize\texttt{$^1$\{zhangtong17,wye,zhanglong418,sjn,zhangsk,zhaowen\}@pku.edu.cn}}\\
\textmd{\normalsize\texttt{$^2$\{yangbaosong.ybs,xingzhang.rxz,liudayiheng.ldyh,zhanhui.zhb\}@alibaba-inc.com}}}

\usepackage{bibentry}
\begin{document}

\maketitle

% \renewcommand{\thefootnote}{\fnsymbol{footnote}}
% \footnotetext[2]{Corresponding authors. This work was done when Tong Zhang was interning at DAMO Academy, Alibaba Group.}

%Infrequent word always plays a crucially semantic role in a sentence. Since the word frequency performs a long-tailed distribution in natural language, rare word learning remains an open challenge in modern neural machine translation (NMT) systems.  
%In this paper, we attempt to understand and tackle this challenge from the representation learning perspective. We first empirically show that the representations of rarer words are distributed in a more compact latent space, raising the difficulty of their prediction. In order to enhance the expressiveness of representations, we propose a novel frequency-aware token-level contrastive learning method. Concretely, according to word frequencies, the hidden state of each decoding step is pushed away from that of other target words with different weights.
%We conduct experiments on widely used NIST Chinese-English and WMT14 English-German translation tasks. Experimental results confirm our hypothesis that the proposed strategy can significantly boost both of translation quality and lexical richness over baseline. Extensive analyses also reveal that, comparing with related adaptive training strategies, the superiority of our method lies in the prediction precision  and the robustness across different translation cases.

\begin{abstract}
Low-frequency word prediction remains a challenge in modern neural machine translation (NMT) systems. 
% Recent adaptive training methods mitigate the under-translation of infrequent words by emphasizing their weights in the overall training objectives. 
Recent adaptive training methods promote the output of infrequent words by emphasizing their weights in the overall training objectives. 
Despite the improved recall of low-frequency words, their prediction precision is unexpectedly hindered by the adaptive objectives. 
Inspired by the observation that low-frequency words form a more compact embedding space, we tackle this challenge from a representation learning perspective. Specifically, we propose a frequency-aware token-level contrastive learning method, in which the hidden state of each decoding step is pushed away from the counterparts of other target words, in a soft contrastive way based on the corresponding word frequencies. 
We conduct experiments on widely used NIST Chinese-English and WMT14 English-German translation tasks. 
Empirical results show that our proposed methods can not only significantly improve the translation quality but also enhance lexical diversity and optimize word representation space. 
Further investigation reveals that, comparing with related adaptive training strategies, the superiority of our method on low-frequency word prediction lies in the robustness of token-level recall across different frequencies without sacrificing precision.

\end{abstract}

\section{Introduction}
\label{sec:intro}

Neural Machine Translation \citep[NMT,][]{sutskever2014sequence,bahdanau2015neural,vaswani2017attention} has made revolutionary advances in the past several years. 
However, the effectiveness of these data-driven NMT systems is heavily reliant on the large-scale training corpus, where the word frequencies demonstrate a long-tailed distribution according to Zipf's Law~\cite{Zipf1949HumanBA}. 
The inherently imbalanced data leads NMT models to commonly prioritize the generation of frequent words while neglect the rare ones.
%The inherently imbalanced data leads NMT models to commonly prioritize the generation of words more frequently appearing in the corpus while neglect the infrequent ones.
% However, since NMT systems heavily depend on the large-scale training corpus, where the word frequencies demonstrate a long-tailed distribution according to Zipf's Law~\cite{zipf}, they commonly prioritize the generation of words more frequently appearing in the corpus. 
Therefore, predicting low-frequency yet semantically rich words remains a bottleneck of current data-driven NMT systems \cite{vanmassenhove-etal-2019-lost}.

% A persisting dilemma for NMT models is that they have trouble in predicting the low-frequency tokens. Whereas, these low-frequency tokens are of vital importance in translations since they often reflect some key semantics of source sentences.

%例子 可视化。低频词为何重要。
% This neglect problem of low-frequency words mainly arises from the extreme imbalance of token distributions. As the Zipf’s Law reveals, neural languages exhibit long-tailed token distributions, i.e. a fraction of tokens are frequent and the rest are low-frequency. However, this token imbalance phenomenon inflict the ``output bias'' on the current data-driven models, where the decision boundary prioritises the generation of words frequently appearing in the corpus. 

\begin{figure}[ht]
	\centering
	\includegraphics[scale=0.71]{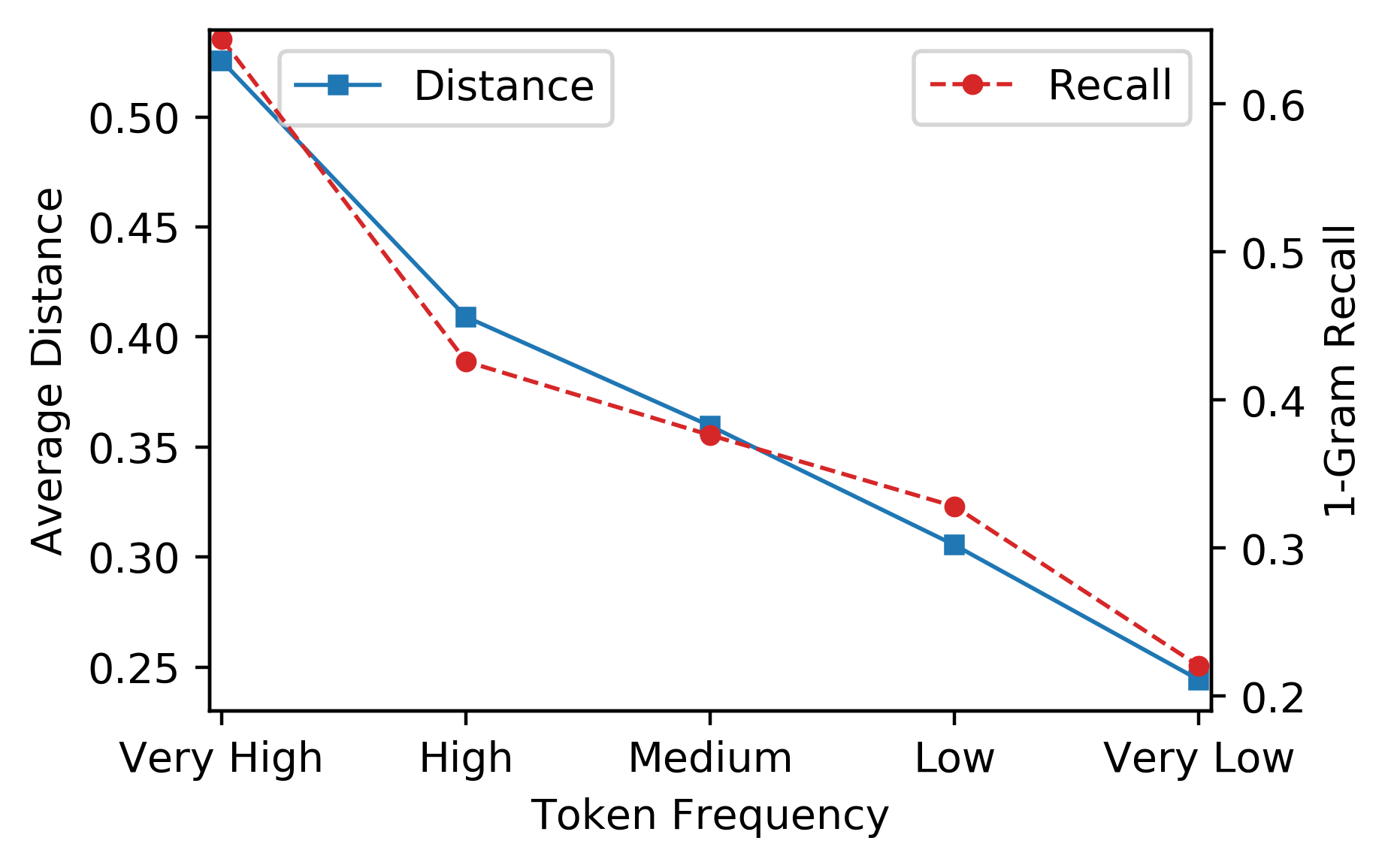}
	\caption{Average token representation distance and 1-gram recall of token buckets with different frequencies on NIST Zh-en test sets and a well-trained vanilla Transformer model. We sort all tokens in target vocabulary based on their frequencies in LDC training set, and divide them into five equal-size buckets. Obviously, the average token distance and 1-gram recall similarly deteriorate with decreasing frequency.}
	%although the target word ``friction" may also be generated instead of copied in practice. }
	\label{fig:example}
\end{figure}

A common practice to facilitate the generation of infrequent words is to smooth the frequency distribution of tokens. For example, it has become a de-facto standard to split words into more fine-grained translation units such as subwords \cite{wu2016google,sennrich2016neural-bpe}.
%characters \cite{luong2016achieving,lee2017fully} or
% A common practice to facilitate the generation of infrequent words is to introduce more fine-grained translation units, such as characters \cite{luong2016achieving,lee2017fully} or subwords \cite{wu2016google,sennrich2016neural-bpe}. 
Despite that, NMT systems still face the token imbalance phenomenon \cite{gu2020token}. 
%As a consequence, NMT systems are commonly misdirected to overproduce the high-frequency tokens while neglect the low-frequency ones too much. 
More recently, some efforts have been dedicated to applying adaptive weights to target tokens in training objectives based on their frequency \cite{gu2020token,Xu2021}. 
By heightening the exposure of low-frequency tokens during training, these models can meliorate the neglect of low-frequency tokens and improve lexical diversity of the translations.
%By heightening the exposure of low-frequency tokens during training, these models can recall more infrequent tokens and improve lexical diversity. 
However, simply promoting low-frequency tokens via loss re-weighting may potentially sacrifice the learning of high-frequency ones
~\cite{gu2020token,DBLP:conf/emnlp/WanYWZCZC20,zhou2020uncertainty}. 
Besides, our further investigation on these methods reveals that generating more unusual tokens comes at the unexpected expense of their prediction precision (Section \ref{sec:1-gram-metric}).

In modern NMT models, the categorical distribution of the predicted word in a decoding step is generated by multiplying the last-layer hidden state by the softmax embedding matrix.\footnote{In the following we will call the softmax embeddings as word embeddings, since sharing them in NMT decoder has been a de-facto standard \cite{iclr2017tying}.} 
%{It has been a de-facto standard to share target word embeddings with softmax parameters in NMT decoder \cite{iclr2017tying}. For simplification, we call them collectively as ``word embeddings''.}
 Therefore, unlike previous explorations, we preliminarily investigate low-frequency word predictions from the perspective of the word representation. As illustrate in Figure~\ref{fig:example}, we divide all the target tokens into several subsets according to their frequencies, and check token-level predictions of each subset based on the vanilla Transformer~\cite{vaswani2017attention}. Our observation is that the average word embedding distance\footnote{L2 distance on normalized word embeddings.} and 1-gram recall\footnote{The 1-gram recall (also known as ROUGE-1 \cite{Lin2004ROUGEAP}) is defined as the number of tokens correctly predicted in output divided by the total tokens in reference.} of these subsets demonstrate a similar downward trend with word frequency decrease.

% Our observation is that the recall of words with different frequencies highly correlates with their distribution characteristics in the embedding space. For example, based on the token-level prediction results of a vanilla Transformer, if we evenly divide all the target tokens into several groups according to their frequencies, the average word distance and 1-gram recall of these groups will demonstrate a similar downward trend with word frequency decrease. Figure ~\ref{fig:example} illustrates this correlation in more detail.

%要不要简单写下，降低了表示能力。类似下面这段
% On the one hand, the expressiveness of representations are greatly limited when they are distributed in a narrow cone in the latent space~\cite{wang2020improving,gao2021simcse}. On the other hand, regarding each decoding step as a multi-class classification task, similar hidden states fail to generate well classification boundaries that are friendly to long-tailed classes (or low-frequency words).    

Another important fact is that the embedding of a ground-truth word and the corresponding hidden state will be pushed together during training to get a more significant likelihood~\cite{gao2019representation}. %我感觉有了这句话，上面那段类似的那句话可以不要，只保留embedding tying。
It inspires us that making the hidden states more diversified could potentially benefit the prediction of low-frequency words. On the one hand, more diversified hidden states could expand word embedding space due to their collaboration in NMT models, which is exactly what we expect given the correlation shown in Figure \ref{fig:example}. On the other hand, regarding each decoding step as a multi-class classification task, more diversified hidden states can generate better classification boundaries that are more friendly to long-tailed classes (or low-frequency words).

% To this end, we propose incorporating contrastive learning into NMT models to improve low-frequency word predictions. Concretely, we propose Frequency-aware Contrastive Learning (FCL), which contrasts the token-level hidden representations of different target tokens within a minibatch. Since long-tailed tokens form a more compact embedding space, we further equip the conventional contrastive learning framework with frequency-aware soft weights.

% Contrast with previous successes on exploiting contrastive learning into NMT~\cite{acl2021multilingual,iclr2021CONTRASTIVE}, we have two main characteristics. Firstly, we propose token-level
% contrastive learning rather than the existed sentence-level ones, thus being able to produce more uniformly distributed hidden states at each decoding step. Secondly, we novelly introduce frequency feature into our framework to boost the translation of low-frequecy tokens. 

To this end, we propose incorporating contrastive learning into NMT models to improve low-frequency word predictions. Our contrastive learning mechanism has two main characteristics. Firstly, unlike previous efforts that contrast at the sentence level~\cite{acl2021multilingual,iclr2021CONTRASTIVE}, we exploit token-level contrast at each decoding step to produce hidden states more uniformly distributed. Secondly, our contrastive learning is frequency-aware.  As long-tailed tokens form a more compact embedding space, we propose to amplify the contrastive effect for relatively low-frequency 
tokens. In particular, for an anchor and one of its negatives, we will apply a soft weight to the corresponding distances based on their frequencies—generally, the lower their frequency, the greater the weight.

We have conducted experiments on Chinese-English and English-German translation tasks. The experimental results demonstrate that our method can significantly outperform the baselines and consistently improve the translation of words with different frequencies, especially rare ones. 

Overall, our contributions are mainly three-fold:

 ~\begin{itemize}

     \item We propose a novel \textbf{F}requency-aware token-level \textbf{C}ontrastive \textbf{L}earning method (FCL) for NMT, providing a new insight of addressing the low-frequency word prediction from the representation learning perspective.

    \item Extensive experiments on Zh-En and En-De translation tasks show that FCL remarkably boosts translation performance, enriches lexical diversity, and improves word representation space.

    \item Compared with previous adaptive training methods, FCL demonstrates the superiority of (1) promoting the output of low-frequency words
    without sacrificing the token-level prediction precision and (2) consistently improving token-level predictions across different frequencies, especially for infrequent words. 
 \end{itemize}

\section{Related Work}

\paragraph{Low-Frequency Word Translation} is a persisting challenge for NMT due to the token imbalance phenomenon.
Conventional researches range from introducing fine-grained translation units \cite{luong2016achieving,lee2017fully}, seeking optimal vocabulary \cite{wu2016google,sennrich2016neural-bpe,gowda2020finding,DBLP:conf/acl/LiuYLZLZZS20}, to incorporating external lexical knowledge \cite{luong2015addressing,arthur2016incorporating,Zhang2021PointDA}.
Recently, some approaches alleviate this problem by well-designed loss function with adaptive weights, in light of the token frequency \cite{gu2020token} or bilingual mutual information \cite{Xu2021}. Inspired by these work, we instead proposed a token-level contrastive learning method and introduce frequency-aware soft weights to adaptively contrast the representations of target words.

\paragraph{Contrastive Learning} has been a widely-used technique to learn representations in both computer vision \cite{Hjelm2019LearningDR,khosla2020supervised,Chen2020ASF} and neural language processing \cite{Logeswaran2018AnEF,fang2020cert,gao2021simcse,DBLP:conf/acl/LinYYLZLHS20}. There are also several recent literatures that attempt to boost machine translation with the effectiveness of contrastive learning. \citet{Yang2019ReducingWO} proposes to reduce word omission by max-margin loss. \citet{acl2021multilingual} learns a universal cross-language representation with a contrastive learning paradigm for multilingual NMT. \citet{iclr2021CONTRASTIVE} adopt a contrastive learning method with perturbed examples to mitigates the exposure bias problem. Different from the previous work that contrasts the sentence-level log-likelihoods or representations, our contrastive learning methods 
pay attention to token-level representations and introduce frequency feature to facilitate rare word generation.

\paragraph{Representation Degeneration} has attracted increasing interest recently \cite{gao2019representation,xu2021leveraging}, which refers that the embedding space learned in language modeling or neural machine translation is squeezed into a narrow cone due to the weight tying trick. 
Recent researches mitigate this issue by performing regularization during training \cite{gao2019representation,wang2020improving}. The distinction between our methods and these work is that our contrastive methods are applied on the hidden representations of the diverse instances in the training corpus rather than directly performing regularizations on the softmax embeddings.

% Moreover, \citet{gao2021simcse} prove that contastive objective can regulate the singular value distribution of sentence-embedding matrix in pretrained language model. 

\begin{figure*}[ht]
	\centering
	\includegraphics[scale=0.485]{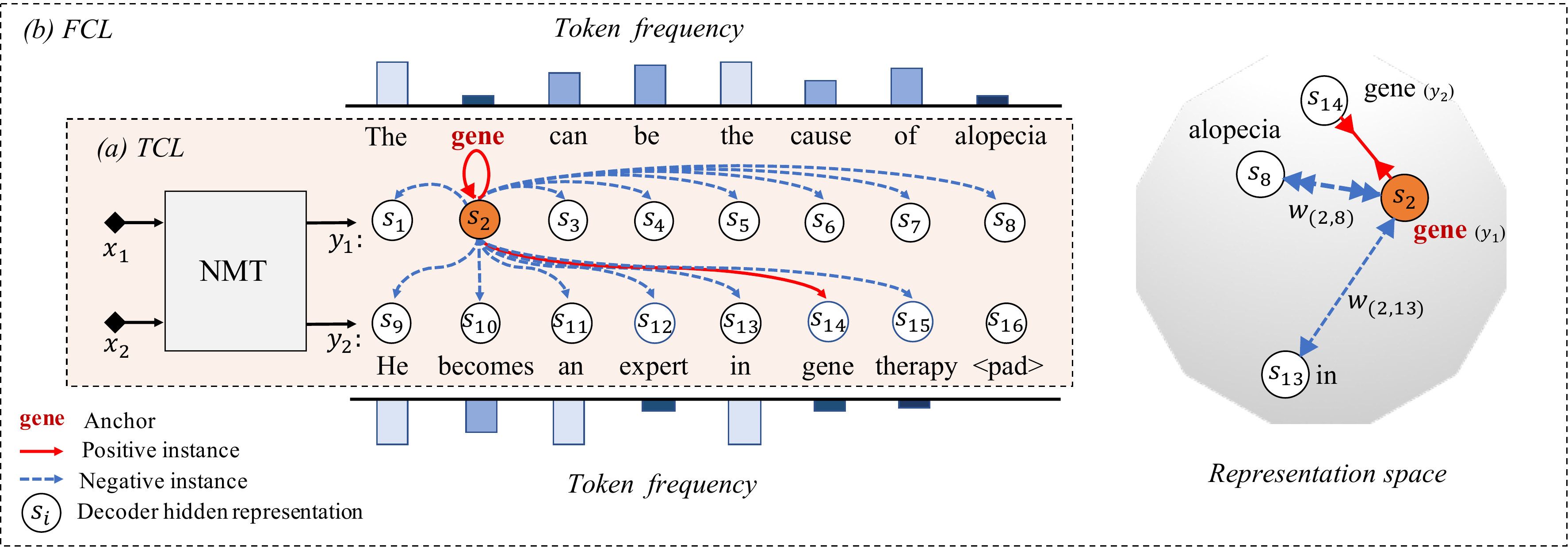}
% 	\includegraphics[scale=0.44,trim =0mm 0mm 0mm 0mm]{graph/Transformer_copy6.pdf}
% 	(a) The proposed token-level contrastive learning method with uniform contrast effect. 
	\caption{
	An example of Token-level Contrastive Learning (TCL) and Frequency-aware Contrastive Learning (FCL). (a) TCL contrasts the token-level hidden representations $s_i$ of the in-batch target tokens. For the anchor ``gene'' in the first sentence $y_1$, there are two sources for its positives, i.e., 
% 	the ``gene'' in $y_2$ and the dropout noise represented by the red self-pointing arrow.
    its counterpart with dropout noise (denoted by the red self-pointing arrow) and the ``gene'' in $y_2$.
	%a supplementary representation of the anchor with dropout noise.
	All other in-batch tokens serve as the negatives.
	(b) FCL further leverages token frequency information to apply frequency-aware soft weights $w(i,j)$ in contrasts. Thus the contrastive effect between relatively infrequent tokens (e.g., ``gene'' and ``alopecia'') is amplified and they can be further pulled 
	apart in the representation space. 
% 	Note that compared with Transformer, TCL and FCL have no extra parameters and require no extra training data, hence demonstrating consistent inference efficiency.
	}
	\label{fig:model}
\end{figure*}

\section{Methodology}
In this section, we mathematically describe the proposed Frequency-aware Contrastive Learning (FCL) method in detail. FCL firstly cast autoregressive neural machine translation as a sequence of classification tasks, and differentiate the hidden representations of different target tokens in Transformer decoder by a token-level contrastive learning method (TCL). 
To facilitate the translation of low-frequency words, we further equip TCL with frequency-aware soft weights, highlighting the classification boundary for the infrequent tokens. An overview of TCL and FCL is illustrated in Figure \ref{fig:model}.
% Note that compared with Transformer, TCL and FCL have no extra parameters and require no extra training data, hence demonstrating consistent inference efficiency.
% We cast autoregressive neural machine translation as a sequence of classification tasks, which predict the next token for the given source sentence and partial target context. 

\subsection{Token-Level Contrastive Learning}
\label{sec:TCL}
In this section we briefly introduce the Token-level Contrastive Learning (TCL) method for NMT. Different from the previous explorations in NMT which contrast the sentence-level representation in the scenario of multilingualism \cite{acl2021multilingual} or adding perturbation \cite{iclr2021CONTRASTIVE}, TCL exploits contrastive learning objectives in the token granularity. Concretely, TCL contrasts the hidden representations of target tokens before softmax classifier. There are two ways in TCL to construct positive instances for each target token: a supervised way to explore the presence of golden label in reference and a supplementary way to take advantage of dropout noise.

\subsubsection{Supervised Contrastive Learning} 
For supervised contrastive learning, the underlying thought is to pulling together the same tokens and pushing apart the different tokens. Inspired by \citet{khosla2020supervised}, we propose a token-level contrastive framework for NMT, which contrasts the in-batch token representations in Transformer decoder. For each target token, we explore the inherent supervised information in the reference to construct the positive samples by the same tokens from the minibatch, and the negative samples are formed by the in-batch tokens different from the anchor. 
In this way, the model can learn effective representation with clear boundaries for the target tokens.

\subsubsection{Supplementary Positives with Dropout Noise} 
In a supervised contrast learning scenario, however, a target token may have no same token in the minibatch due to the token imbalance phenomenon. This limitation is particularly severe when it comes to low-frequency tokens. Thus the supervised contrastive objectives can barely ameliorate the representations of infrequent tokens since in most cases they have no positive instance.
Inspired by \citet{gao2021simcse}, we construct a pseudo positive instance for each anchor by applying independently sampled dropout strategy. By feeding a parallel sentence pair twice into the NMT model and applying different dropout samples, we can obtain a supplementary hidden representation for each target token with dropout noise. In this way, the supplement can serve as the positive instance for the original token representation. Thereby each target token will be assigned at least one positive instance for an effective contrastive learning.

\subsubsection{Formalization} 
The proposed token-level contrastive leaning combines the above two strategies to build the positive instances and use the other in-batch tokens as the negatives.
Formally, given a minibatch with $K$ parallel sentence pairs, $\{{\mathbf x}_k,{\mathbf y}_k\}_{k=1...K}$, which contains a total of $M$ source tokens and $N$ target tokens, the translation probability $p(y_i|\mathbf{y}_{<i},\mathbf{x})$ for the $i^{th}$ in-batch target token $y_i$ is commonly calculated by multiplying the last-layer decoder hidden state $s_i$ by the softmax embedding matrix $\mathbf{W_s}$ in a softmax layer:
\begin{equation}
     p(y_i|\mathbf{y}_{<i},\mathbf{x}) \propto {\rm{exp}}(\mathbf{W_s} \cdot s_i),
     \label{E3}
\end{equation}

In TCL, we feed the inputs to the model twice with independent dropout sample and get two hidden representations $s_i$ and $s'_i$ for a target token $y_i$. For an anchor $s_i$, $s'_i$ serves as the positive, as well as the representations of target tokens that same as $y_i$. The token-level contrastive objective can be formulated as:

% \begin{equation}
% \begin{aligned}
%     L_{cl} =  -\frac{1}{N} \sum_{i=1}^{N} \sum_{p \in P(i)} log\frac{e^{sim(s_i\cdot s_p)/\tau }}{  {\textstyle \sum_{j=1}^{N}} e^{sim(s_i\cdot s_j)/\tau }} \\
%     + {\rm log}\frac{e^{sim(s_i\cdot s'_i)/\tau }}{  {\textstyle \sum_{j=1}^{N}} e^{sim(s_i\cdot s_j)/\tau }},
% \end{aligned}
% \end{equation}
% where $P(i)=\{p \in I \backslash \{i\}:y_p=y_i\}$ is the set of indices of all positive instances.
\vspace{-2ex}
\begin{equation}
    L_{TCL} =  -\frac{1}{N} \sum_{i=1}^{N} \sum_{s_p \in S_p(i)} {\rm log}\frac{e^{sim(s_i\cdot s_p)}}{  {\textstyle \sum_{j=1}^{N}} e^{sim(s_i\cdot s_j)}},
\end{equation}
where $sim(s_i\cdot s_p)$ denotes the cosine similarity between $s_i$ and $s_p$. Here, $S_p(i)= S_{sup}(i) \cup S_{drop}(i)$ is the set of all positive instances for $y_i$. $S_{sup}(i) = \{s_p:p \neq i, p=1...N,y_p=y_i\}$ denotes the positives in supervised contrastive setting, and $S_{drop}(i) = \{s'_i\}$ is the counterpart constructed by dropout noise. 

Finally, the overall training objective combines the traditional NMT objective $L_{MT}$ and the token-level contrastive objective $L_{TCL}$:
\begin{equation}
    L = L_{MT} + \lambda L_{TCL},
\end{equation}
where $\lambda$ is a hyperparameter to balance the effect of TCL.

% $\{s'_i,s_n\},n \neq i, n=1...N,y_n=y_i,$ is the set of all positive instances for $y_i$.
% where $S_p(i)=\{s'_i,s_n|y_n=y_i,n \neq i, n=1...N\} \cup \{s'_i\}$ is the set of indices of all positive instances.

% Contrastive learning aims to learn effective representation by pulling semantically close neighbors together and pushing apart non-neighbors.
\subsection{Frequency-Aware Contrastive Learning}
Token-level contrastive learning treats tokens equally when widening their boundary. However, due to the severe imbalance on token frequency,
a small number of high-frequency tokens make up the majority of token occurrence in a minibatch while most low-frequency words rarely occur in a minibatch. Thus, TCL mainly contrasts the frequent tokens whereas neglects the contrast between the low-frequency tokens. In fact, as illustrated in Section \ref{sec:intro}, the representation of infrequent tokens are more compact and underinformative, which are in more need of amelioration.
Intuitively, in this scenario a better blueprint for contrast is to put more emphasis on the contrast of infrequent tokens.  As a consequence, the model can assign more distinctive representations for infrequent tokens and facilitate the translation of infrequent tokens.
% Intuitively, in this scenario the model should put more emphasis on the contrast between infrequent tokens. As a consequence, the model can assign more distinctive representations for infrequent tokens and facilitate the translation of infrequent tokens.
To this end, we further propose a Frequency-aware Contrast Learning (\textbf{FCL}) method, which utilizes a soft contrastive paradigm to assign frequency-aware soft weights in contrast and thus highlights the contrast of the infrequent tokens.
% between the anchor and its negative samples if they are infrequent.
Formally, for each target token $y_i$, we assign a soft weight $w(i,j)$ for the contrast between the anchor $y_i$ and a negative sample $y_j$. This frequency-aware is determined by the frequencies of both $y_i$ and $y_j$. In FCL, the positives and negatives are built up with the same strategy in token-level contrastive learning. The soft contrastive learning objects can be rewritten as follows:
% We assign a frequency-aware weight for each pair of the anchor $y_i$ and one of its negative sample $y_j$, and the soft contrastive learning objects can be rewritten as follows:
\begin{equation}
    L_{FCL} =  -\frac{1}{N} \sum_{i=1}^{N} \sum_{s_p \in S_p(i)} {\rm log}\frac{e^{sim(s_i\cdot s_p) }}{  {\textstyle \sum_{j=1}^{N}} w(i,j)e^{sim(s_i\cdot s_j) }},
\end{equation}
where $w(i,j)$ is a soft weight in light of frequencies of both the anchor $y_i$ and a negative sample $y_j$. 
The underlying insight is to highlight the contrastive effect for infrequent tokens.
The \textbf{frequency-aware soft weight} $w(i,j)$ is formulated as:
\begin{equation}
\begin{aligned}
    & w(i,j) = {\gamma} f(y_i)f(y_j),\\
   f( y & _i) = 1-\frac{{\rm log( Count}(y_i))}{\underset{j=1...N}{\rm max} {\rm log(Count}(y_j))},
\end{aligned}
\end{equation}
% \begin{equation}
%     w(i,j) = {\gamma} f(y_i)f(y_j),
% \end{equation}
% \begin{equation}
%     f(y_i) = 1-\frac{{\rm count}(y_i)}{\underset{j=1...N}{\rm max} {\rm count}(y_j)}
% \end{equation}
where $f(y_i)$ and $f(y_j)$ are the individual frequency score for $y_i$ and $y_j$, respectively. $\gamma$ is a scale factor for $w(i,j)$. Count$(y_i)$ denotes the word count of $y_i$ in the training set. In our implement, 
% we normalize $f(y_i) = \frac{f(y_i)}{{\rm max}f(\mathbf{y})}$ to ensure $f(y_i)$ ranges in 0 to 1. 
the mean value of frequency-aware weights for all negatives of anchor $y_i$ is normalized to be 1. %删掉？

Accordingly, we weight the traditional NMT objective $L_{MT}$ and the frequency-aware contrastive objective $L_{FCL}$ by a hyperparameter $\lambda$ as follows:
\begin{equation}
    L = L_{MT} + \lambda L_{FCL}.
\end{equation}

% Formally, for each token $y_i$, we first define the frequency-aware weight $w(y_i)$ as:
% \begin{equation}
%     w(y_i) = 1-\frac{f(y_i)}{\underset{j=1...N}{\rm max} f(y_j)},
% \end{equation}
% where $f(y_i)$ denotes the word count of $y_i$ in the training set. In our implement, we normalize $w(y_i) = w(y_i)/{\rm max}w(y_j)$ to ensure
% $w(y_i)$ ranges in 0 to 1. Then we assign a soft weight $W(i,j)$ for the pair of an anchor $y_i$ and a negative sample $y_j$:
% \begin{equation}
%     W(i,j) = w(y_i)*w(y_j)
% \end{equation}

% However, as illustrated in Section \ref{sec:intro}, the representation of low-frequency tokens are extremely converged in comparison to the frequent ones, leading to underperformance on infrequent tokens. 

% Further more, due to the severe imbalance on token frequency, infrequent words rarely appear in a minibatch, which means the contrast are mainly performed between frequent tokens. Intuitively, 

\section{Experimental Settings}
\label{sec:setting}

\begin{table*}[!htpb]
\centering
\begin{tabular}{l||ccccccc||cc}
\hline
\multirow{2}{*}{} & \multicolumn{7}{c||}{Zh-En}   &    \multicolumn{2}{c}{En-De}   \\ 

Systems            & MT02  & MT03  & MT04  & MT05  & MT08 & AVG & $\Delta$ & WMT14&$\Delta$\\ \hline
% \multicolumn{8}{c}{Exsisting NMT systems}  \\ \hline
% \cite{cheng2019robust}     & 46.95 & 47.06 & 46.48 & 47.39 & 46.58 & 37.38 & - \\
% \cite{yang2020improving}   & 44.69 & -     & 46.56 & -     & 46.04 & 37.53 & - \\
% \cite{yan2020multi}        & 47.80 & 47.72 & 46.60 & 48.30 & -     & \textbf{38.70} & - \\\hline
\multicolumn{10}{c}{Baseline NMT systems}               \\ \hline
%Transformer                & 43.60 & 45.71 & 43.21 & 46.52 & 44.24 & 34.47 & ref \\
Transformer         & 47.06 & 46.89 & 47.63 & 45.40 & 35.02 & 44.40 & & 27.84  \\ \hdashline
Focal           & 47.11 & 45.70 & 47.32 & 45.26 & 35.61 & 44.20 & -0.20 &27.91& +0.07 \\
Linear          & 46.84 & 46.27 & 47.26 & 45.62 & 35.59 & 44.32 & -0.08  &28.02 &+0.18 \\
% ~w/ Exp    & 47.54 & 46.99 & 47.52 & 45.57 & 34.99 & 44.52 & +0.12 & 28.17 & +0.33 \\
Exponential             & 46.93 & 47.45 & 47.52 & 46.11 & 36.04 & 44.81 & +0.41 & 28.17 & +0.33 \\
Chi-Square              & 47.14 & 47.15 & 47.68 & 45.46 & 36.15 & 44.72 & +0.32 & 28.31 & +0.47\\
BMI             & 48.05 & 47.11 & 47.64 & 45.97 & 35.93 & 44.94 & +0.54 &28.28 & +0.44\\ 
% \hdashline
CosReg          & 47.11 & 46.98 & 47.72 & 46.60 & 36.66 & 45.01 & +0.61  & 28.38& +0.54\\
% ~w/ SC              & 47.85 & 47.13 & 47.93 & 46.62 & 37.15 & 45.34 & +0.94  & 28.45$^*$\\
\hline
\multicolumn{10}{c}{Our NMT systems}               \\ \hline
% CL                         & 45.80 & 47.44 & 47.80 & 47.86 & 46.43 & 36.76 &  45.23 & +0.83\\
TCL                  & 48.28$^{\dagger\dagger}$ & 47.90$^{\dagger\dagger}$ & 48.31$^{\dagger\dagger}$ & 46.23$^\dagger$~ & 36.67$^{\dagger\dagger}$ &  45.48$^{\dagger\dagger}$ & +1.08 & 28.51$^\dagger$~ & +0.67\\
FCL                 & \textbf{48.95}$^{\dagger\dagger}$ & \textbf{48.63}$^{\dagger\dagger}$ & \textbf{48.38}$^{\dagger\dagger}$ & \textbf{46.82}$^{\dagger\dagger}$ & \textbf{37.00}$^{\dagger\dagger}$ &  \textbf{45.96}$^{\dagger\dagger}$ & \textbf{+1.56} & \textbf{28.65}$^{\dagger\dagger}$ & \textbf{+0.81}\\
% PDC(w/o Dict-Pointer)     & 45.79 & 47.58 & 47.81 & 47.98 & 46.32 & 36.53 &  +1.63\\
% PDC(w/o Tgt-View)          & 45.80 & 47.43 & 47.91 & 48.49 & 46.81 & 36.99 &  +1.91\\ 
% PDC(w/o Src-View)          & 45.97 & 47.42 & 47.90 & 47.92 & 47.07 & 36.81 &  +1.81\\ 
 \hline
\end{tabular}
\caption{Main results on NIST Zh-En and WMT14 En-De tasks. $\Delta$ shows the average BLEU improvements over the test sets compared with Transformer baseline. ``$\dagger$'' and ``$\dagger\dagger$'' indicate the improvement over Transformer is statistically significant ($p<$ 0.05 and $p<$ 0.01, respectively), estimated by bootstrap sampling \cite{Koehn2004StatisticalST}.}
% The results of our models significantly outperform Transformer ($p<0.01$).}
\label{table:main}
\end{table*}

\subsection{Setup}
\paragraph{Data Setting} We evaluate our model on both widely used NIST Chinese-to-Engish (Zh-En) and WMT14 English-German (En-De) translation tasks. 
\begin{itemize}
    \item For Zh-En translation, we use the LDC\footnote{The training set includes LDC2002E18, LDC2003E07, LDC2003E14, Hansards portion of LDC2004T07, LDC2004T08 and LDC2005T06.} corpus as the training set, which consists of 1.25M sentence pairs. We adopt NIST 2006 (MT06) as the validation set and NIST 2002, 2003, 2004, 2005, 2008 datasets as the test sets.
    \item For En-De Translation, the training data contains 4.5M sentence pairs collected from WMT 2014 En-De dataset. We adapt newstest2013 as the validation set and test our model on newstest2014.
\end{itemize}

%which contains 1.6K sentence pairs.
%NIST 2002, 2003, 2004, 2005, 2008 datasets are used for testing.
%which contain 0.9K, 0.9K, 1.8K, 1.1K and 1.4K sentences respectively. 
We adopt Moses tokenizer to deal with English and German sentences, and segment the Chinese sentences with the Stanford Segmentor.\footnote{\url{https://nlp.stanford.edu/}} Following common practices, we employ byte pair encoding \cite{sennrich2016neural-bpe} with 32K merge operations.
% to segment words into subword units. 
%For all tasks we apply BPE \cite{sennrich2016neural-bpe} with 32K merge operations. We use a joint dictionary for English-German translation task while assigning individual vocabularies for Chinese-English translation task. 
% In addition, we remove the examples in datasets where the length of source or target sentence exceeds 100 words.

\paragraph{Implementation Details}
We examine our model based on the advanced Transformer architecture and \emph{base} setting~\cite{vaswani2017attention}.  All the baseline systems and our models are implemented on top of THUMT  toolkit \cite{zhang2017thumt}. 
During training, the dropout rate and label smoothing are set to 0.1.  We employ the Adam optimizer with $\beta_2$ = 0.998.   We use 1 GPU for the NIST Zh-En task and 4 GPUs for WMT14 En-De task. The batch size is 4096 for each GPU. 
% The word embeddings are shared with the softmax matrix in Transformer decoder.
 % with weight tying trick. 
%All these hyper-parameters are tuned on the validation set. 
The other hyper-parameters are the same as the default ``\emph{base}'' configuration in \citet{vaswani2017attention}. 
 The training of each model is early-stopped to maximize BLEU score on the development set. 
 The best single model in validation is used for testing. We use $multi{\rm{-}}bleu.perl$\footnote{https://github.com/moses-smt/mosesdecoder/blob/ master/scripts/generic/multi-bleu.perl} to calculate the case-sensitive BLEU score.% for NIST Zh-En and WMT14 En-De, respectively.

For TCL and FCL, the optimal $\lambda$ for contrastive learning loss is 2.0. The scale factor $\gamma$ in FCL is set to be 1.4. 
% Temperature $\tau$ is the default value 1.0. 
All these hyper-parameters are tuned on the validation set. 

Note that compared with Transformer, TCL and FCL have no extra parameters and require no extra training data, hence demonstrating consistent inference efficiency. Due to the supplementary positives with dropout noise in contrastive objectives, the training speed of FCL is about 1.59$\times$ slower than vanilla Transformer.

\subsection{Baselines}
We re-implement and compare our proposed token-level contrastive learning (TCL) and frequency-aware contrastive learning (FCL) methods with the following baselines:
% \footnote{Note that all these methods need no extra parameters, training data or inference time consumption compared with Transformer.}:

\begin{itemize}

    \item \textbf{Transformer} \cite{vaswani2017attention} is the most widely-used NMT system with self-attention mechanism.
% In order to explore how our model can further improve performance on the basis of BPE, we conduct experiments on Transformers without BPE and Transformer with BPE. All other experiments are with BPE.
    \item \textbf{Focal} \cite{lin2017focal} is a classic adaptive training method proposed for tackling label imbalance problem in object detection.
    In Focal loss, difficult tokens with low prediction probabilities are assigned with higher learning rates. 
    We treat it as a baseline because the low-frequency tokens are intuitively difficult to predict.
    % We treat it as a baseline and assign smaller weights for the tokens with higher prediction probabilities.
    
    \item \textbf{Linear} \cite{jiang2019improving} is an adaptive training method with a linear weight function of word frequency.

    \item \textbf{Exponential} \cite{gu2020token} is an adaptive training method with the exponential weight function. %as shown in Figure \ref{fig:exp&k2}.

    \item \textbf{Chi-Square} \cite{gu2020token} is an adaptive training method that adopts a chi-square distribution as the weight function. The exponential and chi-square weight functions are shown in Figure \ref{subfigure:exp_k2}.

    \item \textbf{BMI} \cite{Xu2021} is a bilingual mutual information based adaptive training objective which estimates the learning difficulty between the source and the target to build the adaptive weighting function.

    \item \textbf{CosReg} \cite{gao2019representation} is a cosine regularization term to maximize the distance between any two word embeddings to mitigate representation degeneration problem. We also treat it as a baseline.
    % \item \textbf{w/ SC} \cite{wang2020improving} is a spectrum control regularization to regulate the singular values of output embeddings.
    
    % \item \textbf{Dropout-based contrastive learning} 
    
    % \item \textbf{Token-level Contrastive learning} 
    
    % \item \textbf{Frequency-aware Contrastive learning} 

\end{itemize}

\section{Experimental Results}

\subsection{Main Results}
Table \ref{table:main} shows the performance of the baseline models and our method variants on NIST Zh-En and WMT En-De translation tasks. We have the following observations.

% As seen, several examined adaptive training methods, e.g., Exponential, Chi-Square and BMI boost NMT on both NIST Zh-En and WMT En-De translation tasks. This is consistent with prior findings that highlighting the loss of low-frequency or unusual words can improve the overall translation quality of NMT models~\cite{gu2020token,Xu2021}. However, the improvement is relatively marginal. We attribute this to the fact that the adaptive training method can not consistently improve the translation quality over all words.  
% The promotion of a type of training samples, to some extent, sacrifices the learning of other cases. We will further explore this problem in Section \ref{sec:1-gram-metrics}. %In our further exploration in Section \ref{sec:1-gram-metrics}, we demonstrate that these method mainly enhance the recall of the high-exposure words, but their prediction precision is compromised.

% Different to adaptive training methods, we approach the problem from the representation learning perspective. Experimental results demonstrate the universal-effectiveness of our method. Specifically, the token-level contrastive learning (TCL) method significantly improves the translation performance on both NIST Zh-En (+1.08 BLEU) and WMT En-De (+0.67 BLEU) tasks over Transformer. By introducing frequency-aware soft weights in contrast objective, the proposed FCL achieves the best performance on both NIST Zh-En and WMT En-De tasks, with gaps to Transformer of 1.56 BLEU and 0.81 BLEU, respectively. 

First, regarding the adaptive training methods, those carefully designed adaptive objectives (e.g., Exponential, Chi-Square, and BMI) achieve slight performance improvement compared with two previous ones (Focal and Linear). As revealed in ~\citet{gu2020token}, Focal and Linear will harm high-frequency token prediction by simply highlighting the loss of low-frequency ones. In fact, our further investigation shows that these newly proposed adaptive objectives alleviate but do not eliminate the negative impact on more frequent tokens, and the increasing weight comes with an unexpected sacrifice in word prediction precision. This is the main reason for their marginal improvements (see more details in Section \ref{sec:1-gram-metric}).

% Second, even as a suboptimal model, our proposed TCL outperforms all adaptive training methods on both translation tasks, verifying the effectiveness of introducing token-level contrastive learning into the decoding process.

Second, even as a suboptimal model, our proposed TCL outperforms all adaptive training methods on both Zh-En and En-De translation tasks. This verifies that expanding softmax embedding latent space can effectively improve the translation quality, which is confirmed again by the results of CosReg.  
%of introducing token-level contrastive learning into the decoding process. In addition, the results of CosReg confirm again that expanding softmax embedding latent space can effectively improve the translation quality. %In comparison to directly performing the regularization on the softmax embeddings, 
Compared with CosReg improving the uniformity using a data-independent regularizer, ours leverage the rich semantic information contained in the training instances to learn superior representations, thus performing a greater improvement. 
%are applied on the diverse hidden representations of the tokens in corpus and achieve a greater improvement. 
%We attribute this to the fact that our methods can leverage the rich semantic information contained in the training instances to learn superior representations rather than improving the uniformity by a data-independent regularizer.

Third, FCL further improves TCL, achieving the best performance and significantly outperform Transformer baseline on NIST Zh-En and WMT En-De tasks. For example, FCL achieves an impressive BLEU improvement of 1.56 over vanilla Transfomer on Zh-En translation. The results clearly demonstrate the merits of incorporating frequency-aware soft weights into contrasting. 

% Finally, the improvement of FCL on WMT En-De task is not as evident as that of NIST Zh-En task (0.81 v.s. 1.56). The possible reason is that the source and target vocabularies are shared in En-De translation. The sharing inherently improves target word embedding space, leaving less room for improvement with contrasting hidden states. Even so, the superiority of FCL and CL over adaptive training methods on WMT En-De task is still notable.

%In addition, the results of CosReg also demonstrate that enhancing the expressiveness of softmax embedding matrix by regulating the uniformity can effectively improve the translation quality. In comparison to directly performing the regularization on the softmax embeddings, our contrastive methods are applied on the diverse hidden representations of the tokens in corpus and achieve a greater improvement. 
%We think the reason is that our methods can leverage the rich semantic information contained in the training instances to learn superior representations rather than improving the uniformity by a data-independent regularizer.

\begin{table}[t]
\centering
\begin{tabular}{l|lll}
\hline
Method          & High & Medium & Low \\ \hline
Transformer & 50.11 & 43.92 & 39.30 \\ \hdashline
Exponential & 49.89\scriptsize\emph{(-0.22)} &44.20\scriptsize\emph{(+0.28)} &40.42\scriptsize\emph{(+1.12)} \\
Chi-Square & 50.13\scriptsize\emph{(+0.02)} &44.18\scriptsize\emph{(+0.26)} &39.95\scriptsize\emph{(+0.65)} \\
BMI &50.35\scriptsize\emph{(+0.24)} &44.02\scriptsize\emph{(+0.10)} &40.44\scriptsize\emph{(+1.14)}\\ \hdashline
TCL & 50.74\scriptsize\emph{(+0.63)} &45.05\scriptsize\emph{(+1.13)} &40.73\scriptsize\emph{(+1.43)} \\
FCL &\textbf{50.95\scriptsize\emph{(+0.84)}} &\textbf{45.42\scriptsize\emph{(+1.50)}} &\textbf{41.30\scriptsize\emph{(+2.00)}}  \\
 \hline
\end{tabular}
\caption{BLEU scores on NIST Zh-En test subsets with different proportions of low-frequency words. ``Low'' denotes the subset in which the target sentences contain more low-frequency words, while ``High'' is the opposite. Our methods yield better translation quality across different subsets.}

\label{tabel:subsets}
\end{table}

\begin{figure*}[t!] 
    \centering
    \subfigure[1-Gram Recall Gap]{
        \begin{minipage}[t]{0.235\textwidth}
            \centering
            \includegraphics[scale = 0.405]{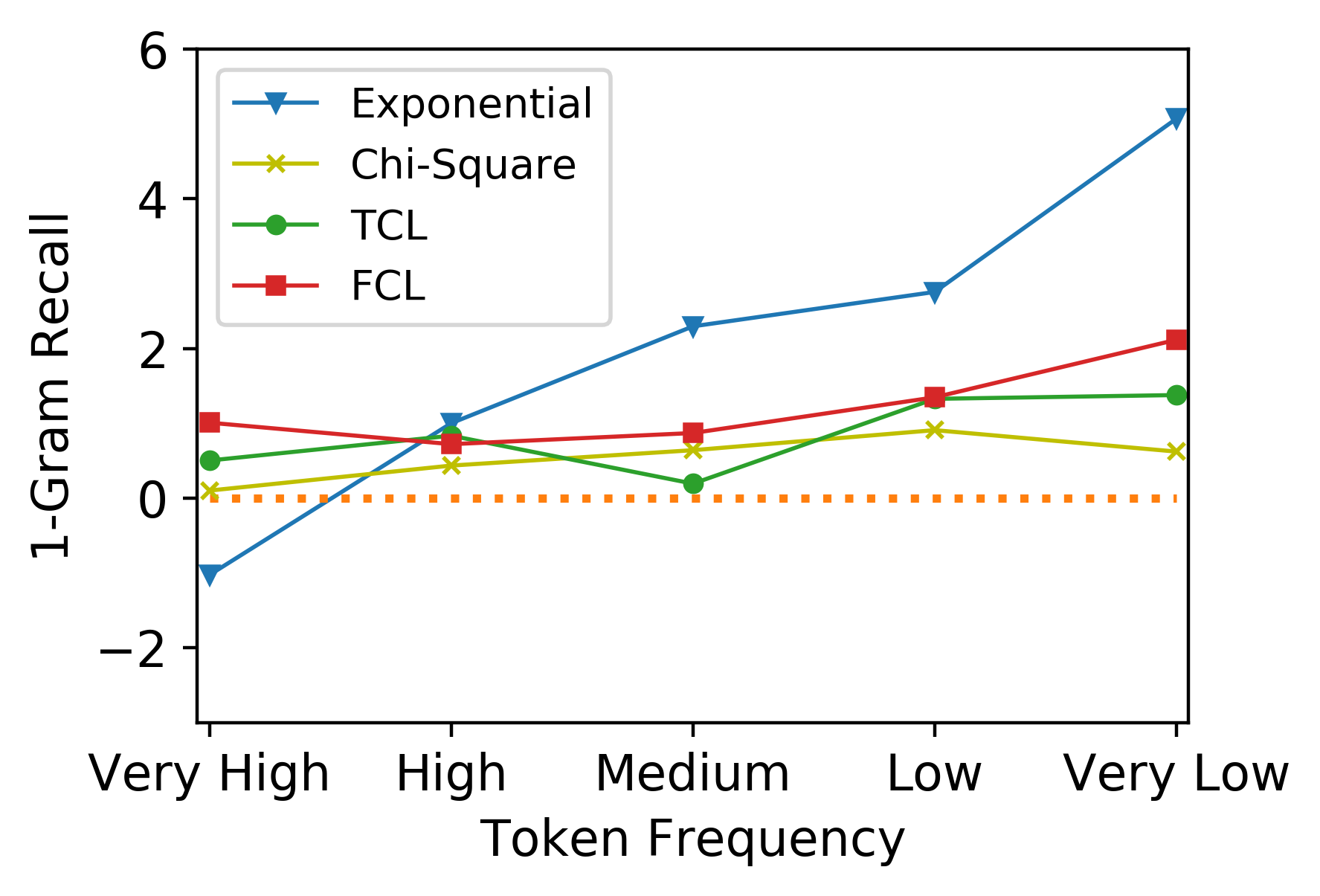}
        \end{minipage}
        \label{Recall_gap} 
    }
    \subfigure[1-Gram Precision Gap]{
        \begin{minipage}[t]{0.235\textwidth}
            \centering
            \includegraphics[scale = 0.405]{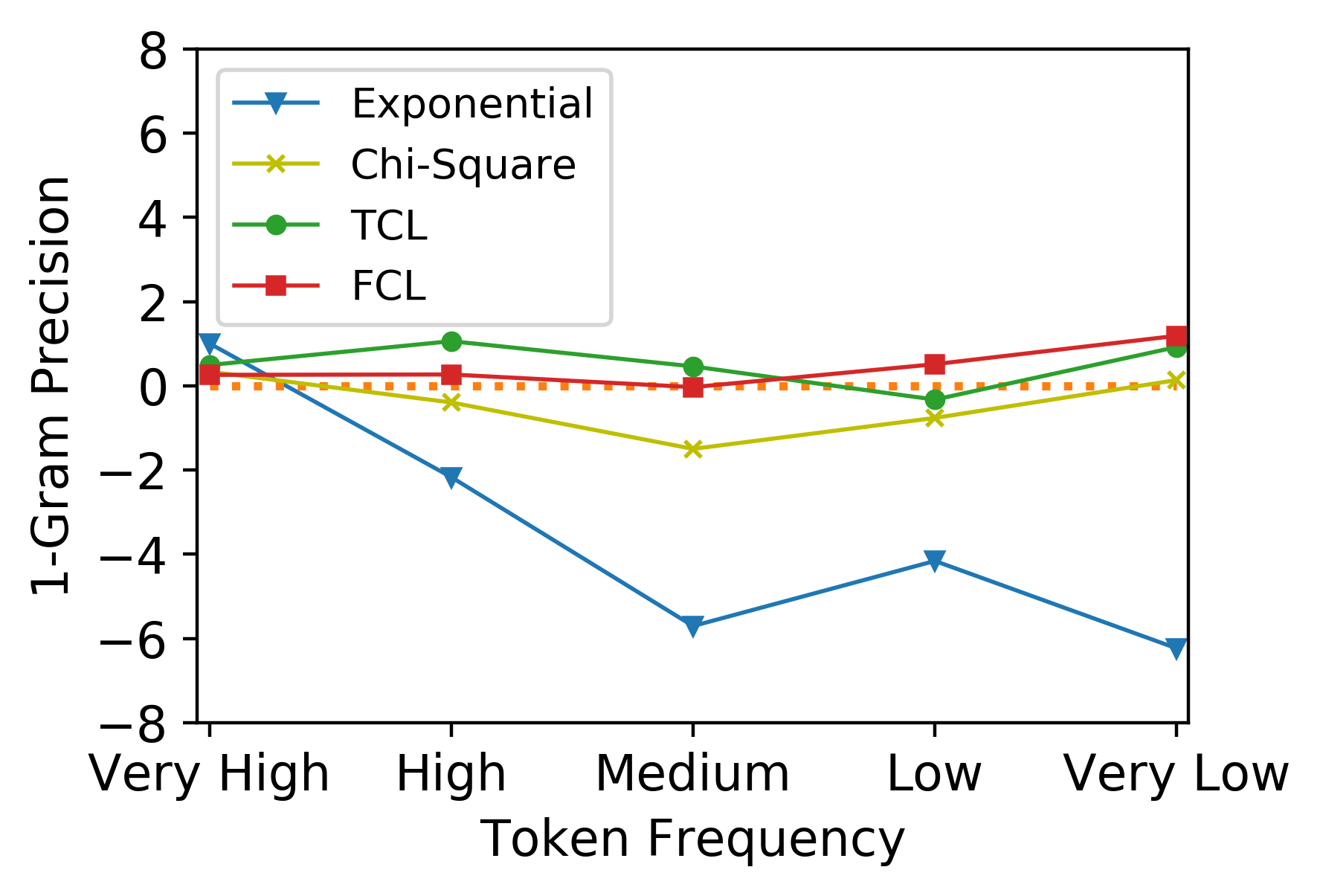}
        \end{minipage}
        \label{Precision_gap} 
    }
    \subfigure[1-Gram F1 Gap]{
        \begin{minipage}[t]{0.235\textwidth}
            \centering
            \includegraphics[scale = 0.405]{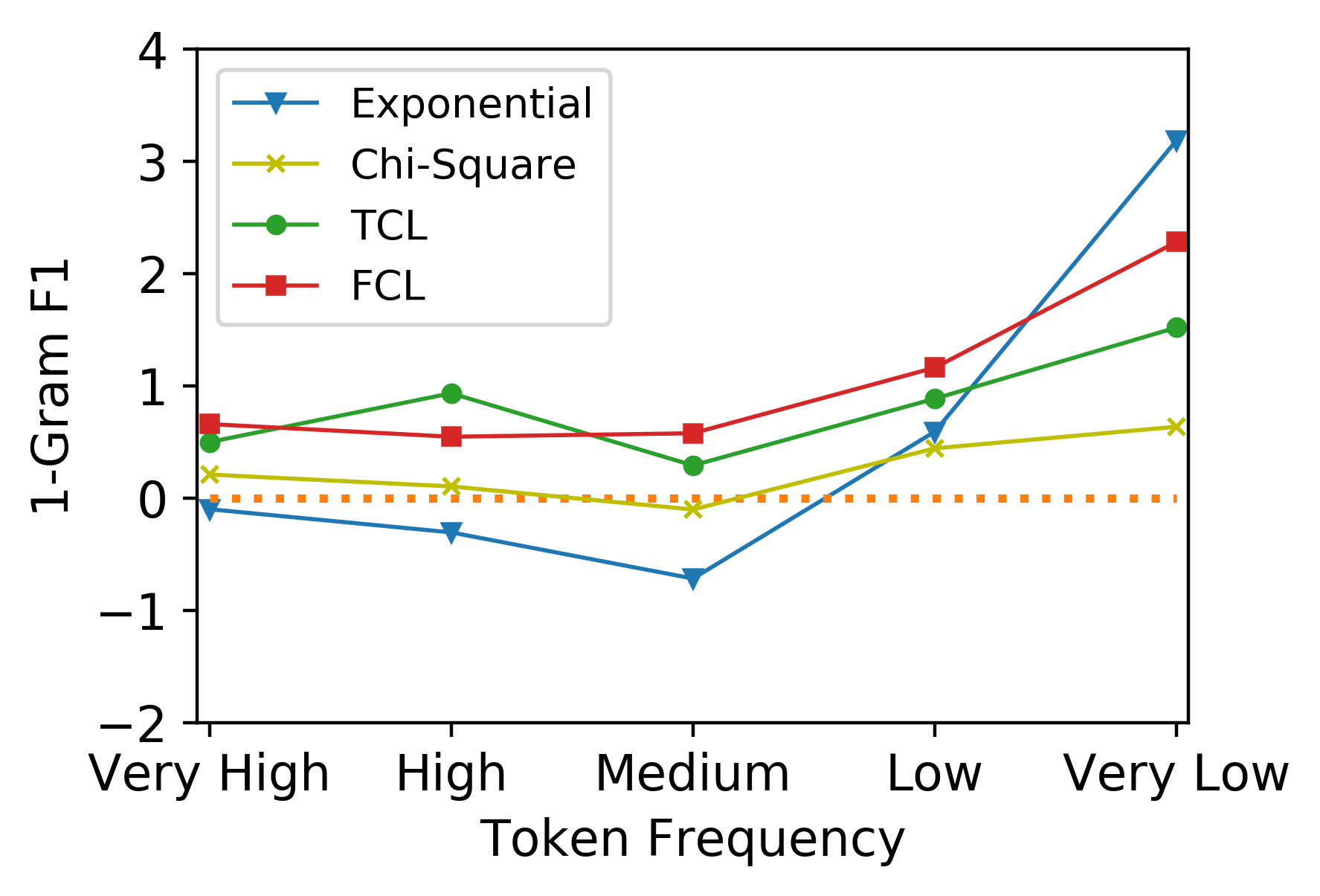}
        \end{minipage}
        \label{F1_gap} 
    }
    \subfigure[Adaptive Weighting Functions]{
        \begin{minipage}[t]{0.235\textwidth}
            \centering
            \includegraphics[scale = 0.405]{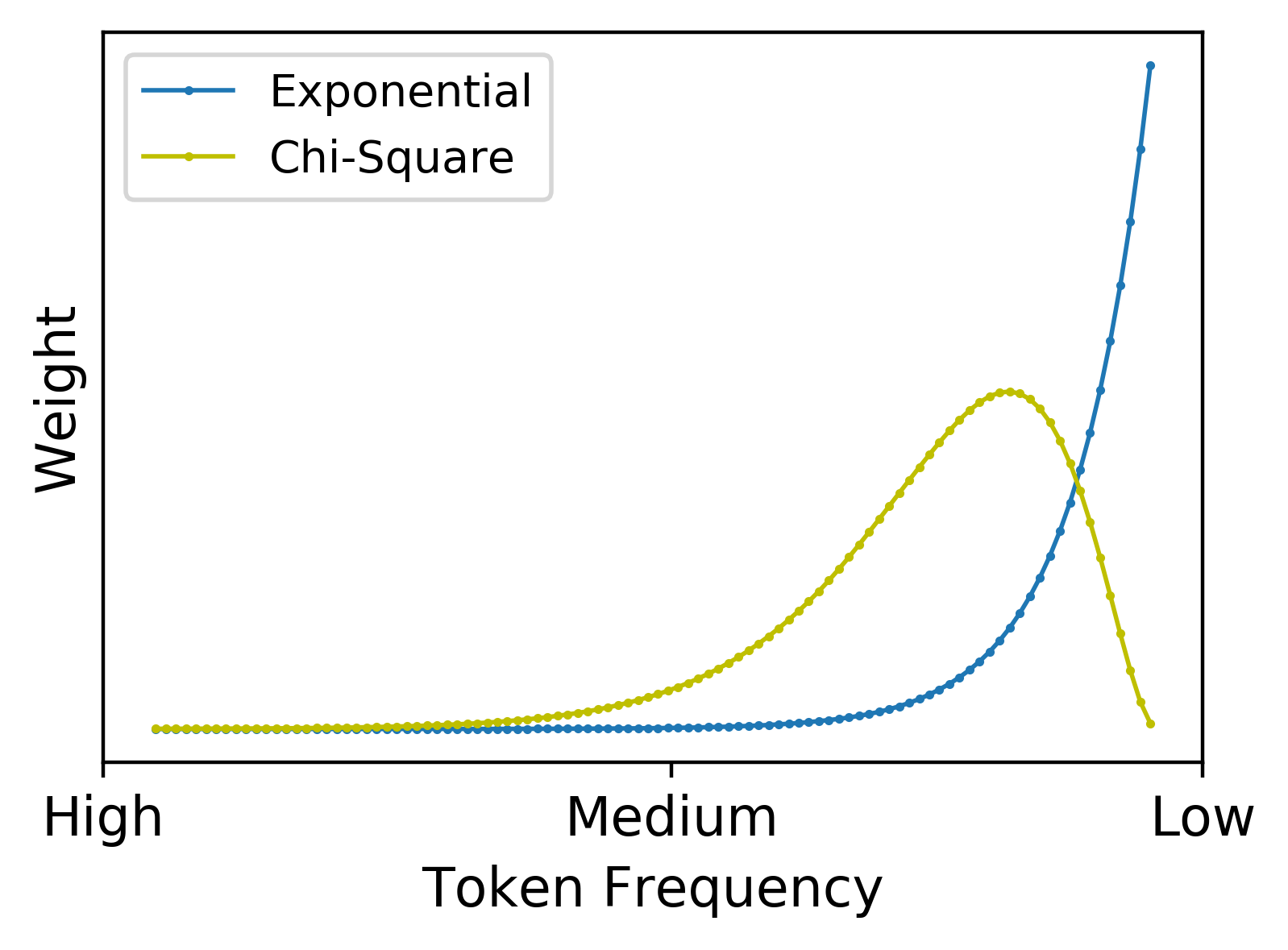}
        \end{minipage}
        \label{subfigure:exp_k2} 
    }
    \caption{(a)/(b)/(c) The 1-Gram Recall/Precision/F1 gaps between various models and transformer. The horizontal dashed line means the Transformer baseline. (d) The Exponential and Chi-Square weighting functions in \citet{gu2020token} with descending token frequency.
    As seen, our methods consistently recall more tokens across different frequencies, in the meanwhile, maintaining the preciseness of prediction.}
        % the different evaluation criteria. (a): The gap on 1-gram recall score;(1-gram accuracy in ..)(b): The gap on 1-gram precision score; (c): The gap on 1-gram F1 score.
    \label{Gap}
\end{figure*}

\subsection{Effects on Translation Quality of Low-Frequency Tokens}
\label{sec:rare-word-quality}

In this section, we investigate the translation quality of low-frequency tokens.\footnote{This experiment and the following ones are all based on Zh-En translation task.} Here we define a target word as a low-frequency word if it appears less than two hundred times in LDC training set\footnote{These words take up the bottom 40\% of the target vocabulary in terms of frequency.}. We then rank the target sentences in NIST test sets by the proportion of low-frequency words and divide them into three subsets: ``High'', ``Middle'', and ``Low''. 

The BLEU scores on the three subsets are shown in Table \ref{tabel:subsets}, from which we can find a similar trend across all methods that the more low-frequency words in a subset, the more notable the performance improvement. Another two more critical observations are that:  (1) FCL and TCL demonstrate their superiority over the adaption training methods across all three subsets of different frequencies. (2) The performance improvements on three subsets we achieved (e.g., 0.84, 1.50, and 2.00 by FCL ) consistently keeps remarkable, while the effects of the adaption training methods on subset ``High'' and ``Middle'' are modest. Note that the Exponential objective even brings a performance degradation on ``High''. These observations verify that our method effectively improves the translation quality of rare tokens, and suggest that improving representation space could be a more robust and systematic way to optimize predictions of diversified-frequency tokens than adaptive training. 

\begin{table}[t]
\centering
\begin{tabular}{l|l l l}
\hline
Method          & MATTR$\uparrow$ & HD-D$\uparrow$ &MTLD$\uparrow$ \\ \hline
Transformer & ~~86.87 & ~86.13 & ~70.96\\  \hdashline
Exponential & ~~87.52 & ~86.86 & ~75.77\\
Chi-Square  & ~~87.16 & ~86.44 & ~71.97\\
BMI         & ~~87.33 & ~86.64 & ~74.05\\ \hdashline
TCL          & ~~87.00 & ~86.25 & ~71.81\\ 
FCL         & ~~87.38 & ~86.71 & ~73.86\\ \hdashline
$Reference$ & ~~88.98 & ~88.23 & ~82.47\\
 \hline
\end{tabular}
\caption{Lexical diversity of translations on NIST test sets. \\ $\uparrow$ means greater value for greater diversity. Both the proposed models and the related studies raise the lexical richness.}
\label{table:lexical_diversity}
\end{table}

\iffalse
\begin{table}[]
\centering
\begin{tabular}{l|lllll}
\hline
            & Seg$_0$     & Seg$_1$     &  Seg$_2$     & Seg$_3$    & Seg$_4$ \\ \hline
Transformer & 87.42 & 7.21 & 3.14 & 1.60 & 0.62 \\\hdashline
Exponential & 85.49 & 7.85 & 3.82 & 1.93 & 0.92 \\
Chi-square  & 87.12 & 7.27 & 3.26 & 1.70 & 0.65 \\
BMI         & 86.93 & 7.39 & 3.27 & 1.69 & 0.71 \\\hdashline
CL          & 87.39 & 7.17 & 3.09 & 1.67 & 0.68 \\
FCL         & 87.23 & 7.22 & 3.19 & 1.65 & 0.71 \\\hdashline
$Reference$   & 84.48 & 7.91 & 3.90 & 2.31 & 1.40 \\\hline
\end{tabular}
\caption{Token distributions with different frequency in references. Seg$_n$ indicates vocabulary segmentation with $n^{th}$ highest average frequency.}
\label{table:frequency}
\end{table}
\fi

\subsection{Effects on Lexical Diversity}

As overly ignoring the infrequent tokens will lead to a lower lexical diversity~\cite{vanmassenhove-etal-2019-lost},  we investigate our methods' effects on lexical diversity following ~\citet{gu2020token} and~\citet{Xu2021}. We statistic three lexical diversity metrics based on the translation results on NIST test sets, including moving-average type-token ratio (MATTR) ~\cite{covington2010cutting}, the approximation of hypergeometric distribution (HD-D) 
% ~\textbf{(Any citation here?)} 
and the measure of textual lexical diversity (MTLD) \cite{mccarthy2010mtld}. The results are reported in Table \ref{table:lexical_diversity}, from which we can observe lexical diversity enhancements brought by FCL and TCL over vanilla Transformer, 
proving the improved tendency of our method to generate low-frequency tokens.

More importantly, we find that Exponential yields the best lexical diversity, though its overall performance improvement in terms of BLEU is far from ours.  This observation inspired us to conduct a more thorough investigation of the token-level predictions, which will be described next.

\begin{figure*}[t!] 
    \centering
    \subfigure[Transformer]{
        \begin{minipage}[t]{0.32\textwidth}
            \centering
            \includegraphics[scale = 0.58]{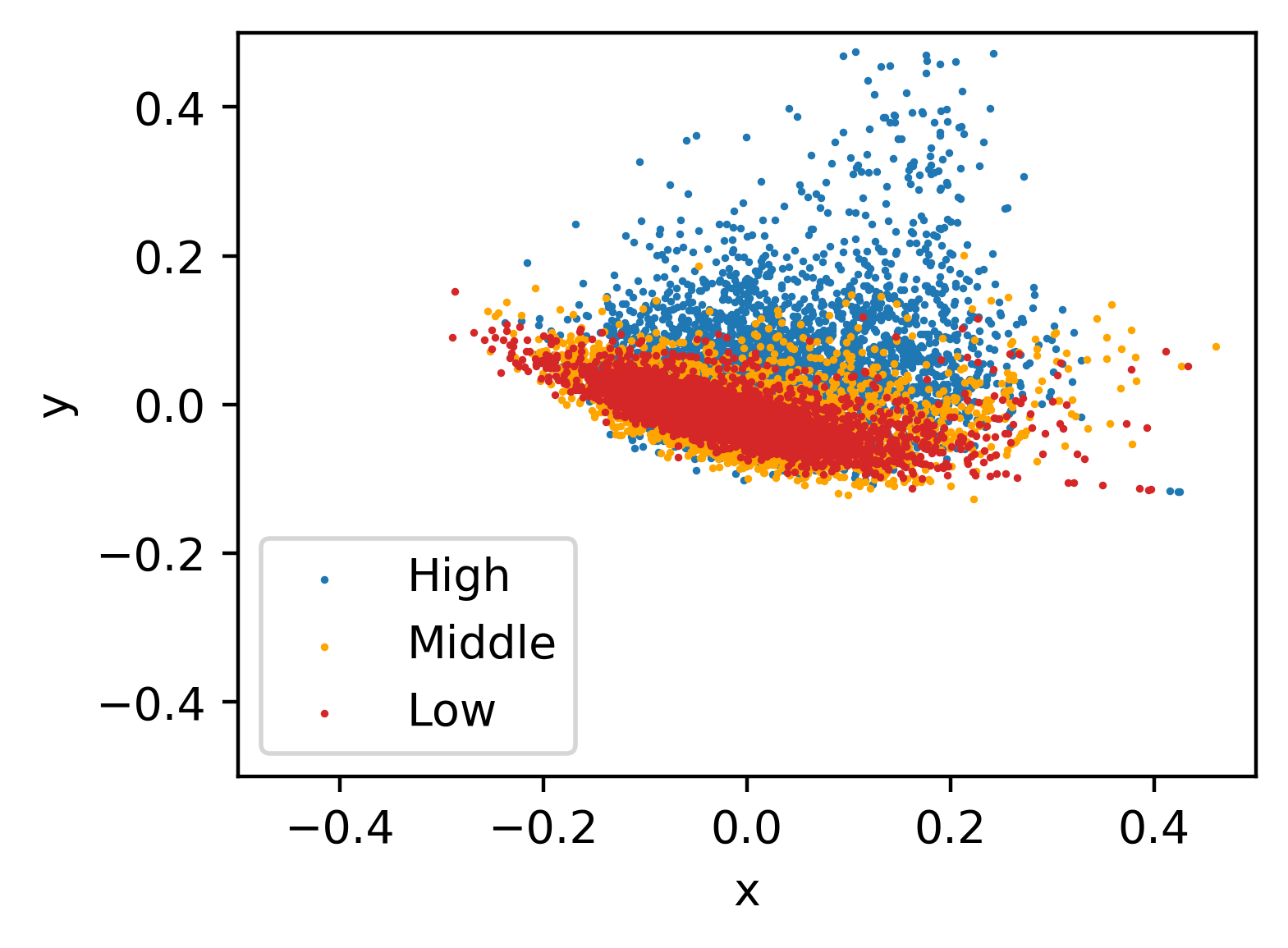}
        \end{minipage}
        \label{Recall_gap} 
    }
    \subfigure[Transformer with TCL]{
        \begin{minipage}[t]{0.32\textwidth}
            \centering
            \includegraphics[scale = 0.58]{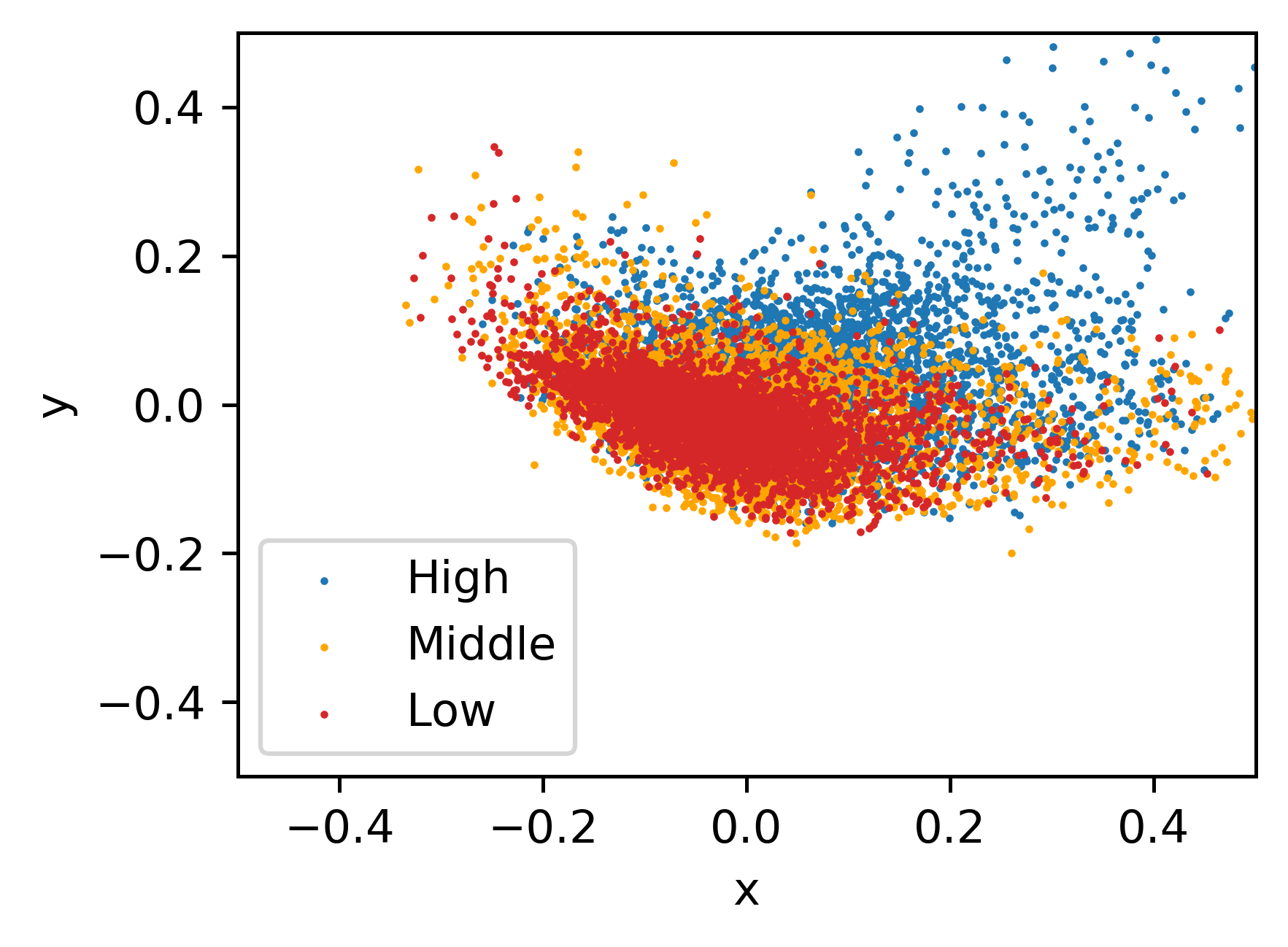}
        \end{minipage}
        \label{Precision_gap} 
    }
    \subfigure[Transformer with FCL]{
        \begin{minipage}[t]{0.32\textwidth}
            \centering
            \includegraphics[scale = 0.58]{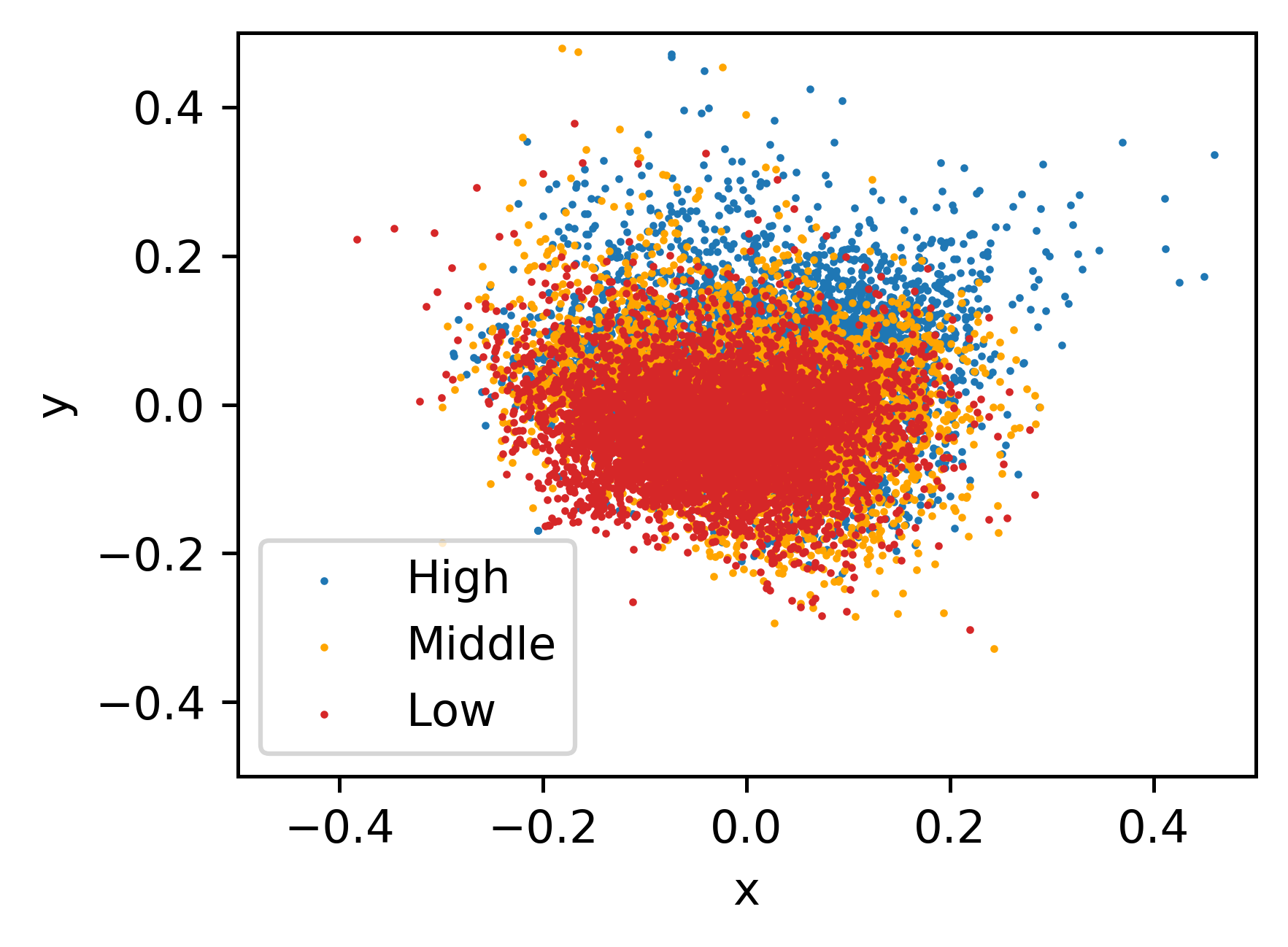}
        \end{minipage}
        \label{F1_gap} 
    }
    \caption{Visualization of word representations (or softmax embeddings) in (a) Transformer, (b) Transformer with TCL, and (c) Transformer with FCL trained on LDC Zh-En dataset. The tokens are evenly divided into three buckets according to the token frequency in the training set. The red dots (``High'') represents the first third of target tokens with high frequency, while the orange and blue ones denote the ``Medium'' and ``Low'' buckets, respectively. In baseline system, the rarer the token, the more serious the problem of representation degeneration is. Obviously, the proposed methods progressively alleviate this problem.}
        % the different evaluation criteria. (a): The gap on 1-gram recall score;(1-gram accuracy in ..)(b): The gap on 1-gram precision score; (c): The gap on 1-gram F1 score.
    \label{embedding}
\end{figure*}

\subsection{Effects on Token-Level Predictions}
\label{sec:1-gram-metric}

Recapping that the three metrics of lexical diversity mainly involve token-level recall, we start by investigating the 1-Gram Recall (or ROUGE-1)
% ~\footnote{also known as ROUGE-1~\cite{Lin2004ROUGEAP}.}
of different methods. First, we evenly divide all the target tokens into five groups according to their frequencies, exactly as we did in Figure 1. The 1-Gram Recall results are then calculated for each group, and the gaps between vanilla Transformer and other methods are illustrated in Figure~\ref{Gap} (a) with descending token frequency.  Regarding the adaptive training methods, we can more clearly see how different token weights affect the token-level predictions by combing Figure~\ref{Gap} (a) and Figure~\ref{Gap} (d) (plots of the Exponential and Chi-Square weighting functions). Here We mainly discuss based on Exponential for convenience, but note that our main findings also apply to Chi-Square. Generally, we find that 1-Gram Recall enhancement is positively related to the token exposure (or weights), roughly explaining why Exponential achieves the best lexical diversity.

Despite the high 1-Gram Recall of Exponential, its unsatisfied overall performance reminds us that the precision of token-level prediction matters. The 1-Gram Precision\footnote{also known as 1-Gram Accuracy in \citet{Feng2020ModelingFA}.} results illustrated in Figure~\ref{Gap} (b) verifies our conjecture.  Different from the trend in Figure \ref{Gap} (a), the gaps of Exponential here looks negatively related to the token exposure. This contrast suggests that though the adaptive training methods generate more low-frequency tokens, the generated ones are more likely to be incorrect. Our FCL and TCL, however, maintain the preciseness of token prediction across different frequencies. The 1-Gram Precision improvements of FCL on low-frequency words (Low and Very Low) are even better than high-frequency ones.

Finally, we investigate the 1-Gram F1,  a more comprehensive metric to evaluate token-level predictions by considering both 1-Gram Recall and Precision. Based on the polylines shown in Figure~\ref{Gap} (C), we can conclude two distinguished merits of our method compared to the adaptive training methods: (1) enhancing low-frequency word recall without sacrificing the prediction precision and (2) improving token-level predictions across different frequencies consistently, which also confirms the observation in Section ~\ref{sec:rare-word-quality}.

\subsection{Effects on Representation Learning}

Since we approach low-frequency word predictions from a representation learning perspective, we finally conduct two experiments to show how our method impacts word representation, though it is not our primary focus.

\begin{table}[t]
\centering
\begin{tabular}{l|l|l|l|l}
\hline
Method          & -Uni $\uparrow$  & Dis $\uparrow$ & $I_1({\rm \mathbf W})\uparrow$ & $I_2({\rm \mathbf W})\downarrow$ \\ \hline
Transformer & 0.2825 & 0.3838 & 0.7446 & 0.0426\\ \hdashline
Exponential & 0.1148 & 0.2431 & 0.7394 & 0.0432\\
Chi-Square & 0.2276 & 0.3442 & 0.7217 & 0.0425\\
BMI & 0.2118 & 0.3190 & 0.7644 & 0.0430\\ \hdashline
TCL  & 0.7024 & 0.5988 & \textbf{0.7903} & 0.0399 \\
FCL & \textbf{0.7490} & \textbf{0.6192} & 0.7652 & \textbf{0.0394}\\
 \hline
\end{tabular}
\caption{The uniformity and isotropy of softmax embeddings on LDC Zh-En machine translation. $\uparrow$ means a positive correlation with the uniformity of the representation and $\downarrow$ represents a negative correlation to the contrary. Our methods can generate more expressive representations. }
\label{uniformity}
\end{table}

Table \ref{uniformity} summarizes the uniformity (Uni)~\cite{DBLP:conf/icml/0001I20}, average distance (Dis) as well as the two isotropy criteria $I_1({\rm \mathbf W})$ and $I_2({\rm \mathbf W})$ \cite{wang2020improving} of the word embedding matrix in NMT systems. Compared with the Transformer baseline and the adaptive training methods, our TCL and FCL substantially improve the measure of uniformity and isotropy, revealing that the target word representations  in our contrastive methods are much more expressive.

To further examine the word representation space, we look at the 2-dimensional visualizations of softmax embeddings by principal component analysis (PCA). One obvious phenomenon in Figure \ref{embedding} (a) is that the tokens with different frequencies lie in different subregions of the representation space. Meanwhile, the embeddings of low-frequency words are squeezed into a more narrow space, which is consistent with the representation degeneration phenomenon proposed by \citet{gao2019representation}. 

 For TCL in Figure \ref{embedding} (b), this problem is alleviated by the token-level contrast of hidden representations, while the differences in word distribution with different word frequencies still persist. As a comparison shown in Figure \ref{embedding} (c), our frequency-aware contrastive learning methods that highlight the contrast of infrequent tokens can produce a more uniform and frequency-robust representation space, leading to the inherent superiority in low-frequency word prediction.

\section{Conclusion}
In this paper, we investigate the problem of low-frequency word prediction in NMT from the representation learning aspect. We propose a novel frequency-aware contrastive learning strategy which can consistently boost translation quality for all words by meliorating word representation space. Via in-depth analyses, our study suggests the following points which may contribute to subsequent researches on this topic: 
1) The softmax representations of rarer words are distributed in a more compact latent space, which correlates the difficulty of their prediction; 2) Differentiating token-level hidden representations of different tokens can meliorates the expressiveness of the representation space and benefit the prediction of infrequent words; 3) Emphasizing the contrast for unusual words with frequency-aware information can further optimize the representation distribution and greatly improve low-frequency word prediction; and
4) In rare word prediction, both recall and precision are essential to be assessed, the proposed 1-Gram F1 can simultaneously consider the two aspects. 

% Previous studies have effectively mitigated the under-translation of infrequent words by assigning adaptive weights on the training objective. However, these methods cannot provide a consistent improvement over words of all frequencies. In this paper, we innovatively tackle this problem from the perspective of representation learning. Our preliminary investigation demonstrate an inherent correlation between 1-gram recall and the aggregation degree of embedding for tokens with different frequencies. From this view, we propose a Frequency-aware Contrastive Leaning method to facilitate the translation of low-frequency words. By contrasting the token-level hidden-representation with frequency-aware soft weights, our FCL not only consistently boosts the translation quality for all words but also enhances the lexical diversity of translations and optimizes word representation space. Further exploration reveals the superiority that FCL can maintaining the precision of prediction while substantially improving the recall of low-frequency words.

\section*{Acknowledgements}
We thank anonymous reviewers for valuable comments. This research was supported by National Key R\&D Program of China under Grant No.2018YFB1403202 and the central government guided local science and technology development fund projects (science and technology innovation base projects) under Grant No.206Z0302G.

\bibliography{ref}

\begin{thebibliography}{41}
\providecommand{\natexlab}[1]{#1}

\bibitem[{Arthur, Neubig, and Nakamura(2016)}]{arthur2016incorporating}
Arthur, P.; Neubig, G.; and Nakamura, S. 2016.
\newblock Incorporating Discrete Translation Lexicons into Neural Machine
  Translation.
\newblock In \emph{EMNLP}.

\bibitem[{Bahdanau, Cho, and Bengio(2015)}]{bahdanau2015neural}
Bahdanau, D.; Cho, K.; and Bengio, Y. 2015.
\newblock Neural Machine Translation by Jointly Learning to Align and
  Translate.
\newblock In \emph{ICLR}.

\bibitem[{Chen et~al.(2020)Chen, Kornblith, Norouzi, and Hinton}]{Chen2020ASF}
Chen, T.; Kornblith, S.; Norouzi, M.; and Hinton, G.~E. 2020.
\newblock A Simple Framework for Contrastive Learning of Visual
  Representations.
\newblock \emph{ICML}.

\bibitem[{Covington and McFall(2010)}]{covington2010cutting}
Covington, M.~A.; and McFall, J.~D. 2010.
\newblock Cutting the Gordian Knot: The Moving-Average Type--Token Ratio
  (MATTR).
\newblock \emph{Journal of quantitative linguistics}, 17(2): 94--100.

\bibitem[{Fang et~al.(2020)Fang, Wang, Zhou, Ding, and Xie}]{fang2020cert}
Fang, H.; Wang, S.; Zhou, M.; Ding, J.; and Xie, P. 2020.
\newblock CERT: Contrastive Self-Supervised Learning for Language
  Understanding.
\newblock \emph{arXiv preprint arXiv:2005.12766}.

\bibitem[{Feng et~al.(2020)Feng, Xie, Gu, Shao, Zhang, Yang, and
  Yu}]{Feng2020ModelingFA}
Feng, Y.; Xie, W.; Gu, S.; Shao, C.; Zhang, W.; Yang, Z.; and Yu, D. 2020.
\newblock Modeling Fluency and Faithfulness for Diverse Neural Machine
  Translation.
\newblock In \emph{AAAI}.

\bibitem[{Gao et~al.(2019)Gao, He, Tan, Qin, Wang, and
  Liu}]{gao2019representation}
Gao, J.; He, D.; Tan, X.; Qin, T.; Wang, L.; and Liu, T.-Y. 2019.
\newblock Representation Degeneration Problem in Training Natural Language
  Generation Models.
\newblock In \emph{ICLR}.

\bibitem[{Gao, Yao, and Chen(2021)}]{gao2021simcse}
Gao, T.; Yao, X.; and Chen, D. 2021.
\newblock SimCSE: Simple Contrastive Learning of Sentence Embeddings.
\newblock \emph{arXiv preprint arXiv:2104.08821}.

\bibitem[{Gowda and May(2020)}]{gowda2020finding}
Gowda, T.; and May, J. 2020.
\newblock Finding the Optimal Vocabulary Size for Neural Machine Translation.
\newblock In \emph{EMNLP: Findings}, 3955--3964.

\bibitem[{Gu et~al.(2020)Gu, Zhang, Meng, Feng, Xie, Zhou, and
  Yu}]{gu2020token}
Gu, S.; Zhang, J.; Meng, F.; Feng, Y.; Xie, W.; Zhou, J.; and Yu, D. 2020.
\newblock Token-Level Adaptive Training for Neural Machine Translation.
\newblock In \emph{EMNLP}, 1035--1046.

\bibitem[{Hakan, Khashayar, and Richard(2017)}]{iclr2017tying}
Hakan, I.; Khashayar, K.; and Richard, S. 2017.
\newblock Tying Word Vectors and Word Classifiers: A Loss Framework for
  Language Modeling.
\newblock In \emph{ICLR}.

\bibitem[{Hjelm et~al.(2019)Hjelm, Fedorov, Lavoie-Marchildon, Grewal,
  Trischler, and Bengio}]{Hjelm2019LearningDR}
Hjelm, R.~D.; Fedorov, A.; Lavoie-Marchildon, S.; Grewal, K.; Trischler, A.;
  and Bengio, Y. 2019.
\newblock Learning Deep Representations by Mutual Information Estimation and
  Maximization.
\newblock \emph{ICLR}.

\bibitem[{Jiang et~al.(2019)Jiang, Ren, Monz, and
  de~Rijke}]{jiang2019improving}
Jiang, S.; Ren, P.; Monz, C.; and de~Rijke, M. 2019.
\newblock Improving Neural Response Diversity with Frequency-Aware
  Cross-Entropy Loss.
\newblock In \emph{The World Wide Web Conference}, 2879--2885.

\bibitem[{Khosla et~al.(2020)Khosla, Teterwak, Wang, Sarna, Tian, Isola,
  Maschinot, Liu, and Krishnan}]{khosla2020supervised}
Khosla, P.; Teterwak, P.; Wang, C.; Sarna, A.; Tian, Y.; Isola, P.; Maschinot,
  A.; Liu, C.; and Krishnan, D. 2020.
\newblock Supervised Contrastive Learning.
\newblock \emph{NeurIPS}.

\bibitem[{Koehn(2004)}]{Koehn2004StatisticalST}
Koehn, P. 2004.
\newblock Statistical Significance Tests for Machine Translation Evaluation.
\newblock In \emph{EMNLP}.

\bibitem[{Lee, Cho, and Hofmann(2017)}]{lee2017fully}
Lee, J.; Cho, K.; and Hofmann, T. 2017.
\newblock Fully Character-Level Neural Machine Translation without Explicit
  Segmentation.
\newblock \emph{Transactions of the Association for Computational Linguistics},
  5: 365--378.

\bibitem[{Lee, Lee, and Hwang(2021)}]{iclr2021CONTRASTIVE}
Lee, S.; Lee, D.~B.; and Hwang, S.~J. 2021.
\newblock Contrastive Learning with Adversarial Perturbations for Conditional
  Text Generation.
\newblock In \emph{ICLR}.

\bibitem[{Lin(2004)}]{Lin2004ROUGEAP}
Lin, C.-Y. 2004.
\newblock ROUGE: A Package for Automatic Evaluation of Summaries.
\newblock In \emph{ACL}.

\bibitem[{Lin et~al.(2021)Lin, Yao, Yang, Liu, Zhang, Luo, Huang, and
  Su}]{DBLP:conf/acl/LinYYLZLHS20}
Lin, H.; Yao, L.; Yang, B.; Liu, D.; Zhang, H.; Luo, W.; Huang, D.; and Su, J.
  2021.
\newblock Towards User-Driven Neural Machine Translation.
\newblock In \emph{ACL}.

\bibitem[{Lin et~al.(2017)Lin, Goyal, Girshick, He, and
  Doll{\'a}r}]{lin2017focal}
Lin, T.-Y.; Goyal, P.; Girshick, R.; He, K.; and Doll{\'a}r, P. 2017.
\newblock Focal Loss for Dense Object Detection.
\newblock In \emph{Proceedings of the IEEE international conference on computer
  vision}, 2980--2988.

\bibitem[{Liu et~al.(2021)Liu, Yang, Liu, Zhang, Luo, Zhang, Zhang, and
  Su}]{DBLP:conf/acl/LiuYLZLZZS20}
Liu, X.; Yang, B.; Liu, D.; Zhang, H.; Luo, W.; Zhang, M.; Zhang, H.; and Su,
  J. 2021.
\newblock Bridging Subword Gaps in Pretrain-Finetune Paradigm for Natural
  Language Generation.
\newblock In \emph{ACL}.

\bibitem[{Logeswaran and Lee(2018)}]{Logeswaran2018AnEF}
Logeswaran, L.; and Lee, H. 2018.
\newblock An Efficient Framework for Learning Sentence Representations.
\newblock \emph{ICLR}.

\bibitem[{Luong and Manning(2016)}]{luong2016achieving}
Luong, M.-T.; and Manning, C.~D. 2016.
\newblock Achieving Open Vocabulary Neural Machine Translation with Hybrid
  Word-Character Models.
\newblock \emph{arXiv preprint arXiv:1604.00788}.

\bibitem[{Luong et~al.(2015)Luong, Sutskever, Le, Vinyals, and
  Zaremba}]{luong2015addressing}
Luong, M.-T.; Sutskever, I.; Le, Q.; Vinyals, O.; and Zaremba, W. 2015.
\newblock Addressing the Rare Word Problem in Neural Machine Translation.
\newblock In \emph{ACL}.

\bibitem[{McCarthy and Jarvis(2010)}]{mccarthy2010mtld}
McCarthy, P.~M.; and Jarvis, S. 2010.
\newblock MTLD, vocd-D, and HD-D: A Validation Study of Sophisticated
  Approaches to Lexical Diversity Assessment.
\newblock \emph{Behavior research methods}, 42(2): 381--392.

\bibitem[{Pan et~al.(2021)Pan, Wang, Wu, and Li}]{acl2021multilingual}
Pan, X.; Wang, M.; Wu, L.; and Li, L. 2021.
\newblock Contrastive Learning for Many-to-many Multilingual Neural Machine
  Translation.
\newblock In Zong, C.; Xia, F.; Li, W.; and Navigli, R., eds.,
  \emph{ACL/IJCNLP}.

\bibitem[{Sennrich, Haddow, and Birch(2016)}]{sennrich2016neural-bpe}
Sennrich, R.; Haddow, B.; and Birch, A. 2016.
\newblock Neural Machine Translation of Rare Words with Subword Units.
\newblock In \emph{ACL}.

\bibitem[{Sutskever, Vinyals, and Le(2014)}]{sutskever2014sequence}
Sutskever, I.; Vinyals, O.; and Le, Q.~V. 2014.
\newblock Sequence to Sequence Learning with Neural Networks.
\newblock \emph{NeurIPS}.

\bibitem[{Vanmassenhove, Shterionov, and
  Way(2019)}]{vanmassenhove-etal-2019-lost}
Vanmassenhove, E.; Shterionov, D.; and Way, A. 2019.
\newblock Lost in Translation: Loss and Decay of Linguistic Richness in Machine
  Translation.
\newblock In \emph{Proceedings of Machine Translation Summit XVII: Research
  Track}, 222--232. Dublin, Ireland: European Association for Machine
  Translation.

\bibitem[{Vaswani et~al.(2017)Vaswani, Shazeer, Parmar, Uszkoreit, Jones,
  Gomez, Kaiser, and Polosukhin}]{vaswani2017attention}
Vaswani, A.; Shazeer, N.; Parmar, N.; Uszkoreit, J.; Jones, L.; Gomez, A.~N.;
  Kaiser, {\L}.; and Polosukhin, I. 2017.
\newblock Attention is All You Need.
\newblock In \emph{NeurIPS}.

\bibitem[{Wan et~al.(2020)Wan, Yang, Wong, Zhou, Chao, Zhang, and
  Chen}]{DBLP:conf/emnlp/WanYWZCZC20}
Wan, Y.; Yang, B.; Wong, D.~F.; Zhou, Y.; Chao, L.~S.; Zhang, H.; and Chen, B.
  2020.
\newblock Self-Paced Learning for Neural Machine Translation.
\newblock In \emph{EMNLP}.

\bibitem[{Wang et~al.(2020)Wang, Huang, Huang, Hu, Wang, and
  Gu}]{wang2020improving}
Wang, L.; Huang, J.; Huang, K.; Hu, Z.; Wang, G.; and Gu, Q. 2020.
\newblock Improving Neural Language Generation with Spectrum Control.
\newblock In \emph{ICLR}.

\bibitem[{Wang and Isola(2020)}]{DBLP:conf/icml/0001I20}
Wang, T.; and Isola, P. 2020.
\newblock Understanding Contrastive Representation Learning through Alignment
  and Uniformity on the Hypersphere.
\newblock In \emph{ICML}.

\bibitem[{Wu et~al.(2016)Wu, Schuster, Chen, Le, Norouzi, Macherey, Krikun,
  Cao, Gao, Macherey et~al.}]{wu2016google}
Wu, Y.; Schuster, M.; Chen, Z.; Le, Q.~V.; Norouzi, M.; Macherey, W.; Krikun,
  M.; Cao, Y.; Gao, Q.; Macherey, K.; et~al. 2016.
\newblock Google's Neural Machine Translation System: Bridging the Gap Between
  Human and Machine Translation.
\newblock \emph{arXiv preprint arXiv:1609.08144}.

\bibitem[{Xu et~al.(2021{\natexlab{a}})Xu, Yang, Lv, Bi, Liu, and
  Zhang}]{xu2021leveraging}
Xu, L.; Yang, B.; Lv, X.; Bi, T.; Liu, D.; and Zhang, H. 2021{\natexlab{a}}.
\newblock Leveraging Advantages of Interactive and Non-Interactive Models for
  Vector-Based Cross-Lingual Information Retrieval.
\newblock \emph{arXiv preprint arXiv:2111.01992}.

\bibitem[{Xu et~al.(2021{\natexlab{b}})Xu, Liu, Meng, Zhang, Xu, and
  Zhou}]{Xu2021}
Xu, Y.; Liu, Y.; Meng, F.; Zhang, J.; Xu, J.; and Zhou, J. 2021{\natexlab{b}}.
\newblock Bilingual Mutual Information Based Adaptive Training for Neural
  Machine Translation.
\newblock In Zong, C.; Xia, F.; Li, W.; and Navigli, R., eds.,
  \emph{ACL/IJCNLP}.

\bibitem[{Yang et~al.(2019)Yang, Cheng, Liu, and Sun}]{Yang2019ReducingWO}
Yang, Z.; Cheng, Y.; Liu, Y.; and Sun, M. 2019.
\newblock Reducing Word Omission Errors in Neural Machine Translation: A
  Contrastive Learning Approach.
\newblock In \emph{ACL}.

\bibitem[{Zhang et~al.(2017)Zhang, Ding, Shen, Cheng, Sun, Luan, and
  Liu}]{zhang2017thumt}
Zhang, J.; Ding, Y.; Shen, S.; Cheng, Y.; Sun, M.; Luan, H.; and Liu, Y. 2017.
\newblock Thumt: An open source toolkit for neural machine translation.
\newblock \emph{arXiv preprint arXiv:1706.06415}.

\bibitem[{Zhang et~al.(2021)Zhang, Zhang, Ye, Li, Sun, Zhu, Zhao, and
  Zhang}]{Zhang2021PointDA}
Zhang, T.; Zhang, L.; Ye, W.; Li, B.; Sun, J.; Zhu, X.; Zhao, W.; and Zhang, S.
  2021.
\newblock Point, Disambiguate and Copy: Incorporating Bilingual Dictionaries
  for Neural Machine Translation.
\newblock In \emph{ACL/IJCNLP}.

\bibitem[{Zhou et~al.(2020)Zhou, Yang, Wong, Wan, and
  Chao}]{zhou2020uncertainty}
Zhou, Y.; Yang, B.; Wong, D.~F.; Wan, Y.; and Chao, L.~S. 2020.
\newblock Uncertainty-aware curriculum learning for neural machine translation.
\newblock In \emph{ACL}.

\bibitem[{Zipf(1949)}]{Zipf1949HumanBA}
Zipf, G. 1949.
\newblock Human Behavior and The Principle of Least Effort.

\end{thebibliography}

\end{document}